\documentclass[lettersize, journal]{IEEEtran}
\usepackage[noadjust]{cite}
\usepackage{amsmath,amsfonts}

\DeclareMathOperator*{\argmin}{arg\,min}
\usepackage{array}
\usepackage{bm}
\usepackage[caption=false,font=normalsize,labelfont=sf,textfont=sf]{subfig}
\usepackage{textcomp}
\usepackage[dvipsnames]{xcolor}
\usepackage{booktabs}
\usepackage{stfloats}
\usepackage{url}
\usepackage{verbatim}
\usepackage{graphicx}
\usepackage[export]{adjustbox}
\usepackage[ruled,linesnumbered,commentsnumbered,vlined]{algorithm2e}
\usepackage{amsthm}
\usepackage{multirow}
\newtheorem*{remark}{Remark}

\usepackage{booktabs,ragged2e}
\usepackage[flushleft]{threeparttable}

\begin{document}
\title{CSDO: Enhancing Efficiency and Success in Large-Scale Multi-Vehicle Trajectory Planning}


\author{Yibin Yang$^{1}$, Shaobing Xu$^{1}$, Xintao Yan$^{2}$, Junkai Jiang$^{1}$, Jianqiang Wang$^{1}$, Heye Huang$^{1}$
\thanks{Manuscript received: March, 18, 2024; Revised June, 29, 2024; Accepted July, 26, 2024.}
\thanks{This paper was recommended for publication by Editor M. Ani Hsieh upon evaluation of the Associate Editor and Reviewers' comments.
This work was supported in part by NSFC (52221005) and the Key Project (52131201). 
\textit{(Corresponding author: Heye Huang.)}
}


\thanks{$^{1}$Y. Yang, J. Jiang, S. Xu, J. Wang and H. Huang are with the School of Vehicle and Mobility, Tsinghua University, Beijing 100084, China. (email: {yyb19,hhy18,jiangjk21}@mails.tsinghua.edu.cn; {shaobxu, wjqlws}@tsinghua.edu.cn).}
\thanks{
$^{2}$X. Yan is with the Department of Civil and Environmental Engineering, University of Michigan, Ann Arbor, MI 48109 USA. (email: xintaoy@umich.edu)
}
\thanks{Digital Object Identifier (DOI): see top of this page.}
}

\markboth{IEEE Robotics and Automation Letters. Preprint Version. Accepted July, 2024}
{Yang \MakeLowercase{\textit{et al.}}: CSDO_LargeScaleMVTP}

\IEEEpubid{0000--0000/00\$00.00~\copyright~2021 IEEE}

\maketitle

\begin{abstract}

This paper presents an efficient algorithm, naming Centralized Searching and Decentralized Optimization (CSDO), to find feasible solution for large-scale Multi-Vehicle Trajectory Planning (MVTP) problem. 
Due to the intractable growth of non-convex constraints with the number of agents, exploring various homotopy classes that imply different convex domains, is crucial for finding a feasible solution.
However, existing methods struggle to explore various homotopy classes efficiently due to combining it with time-consuming, precise trajectory solution finding.
CSDO, addresses this limitation by separating them into different levels and integrating an efficient Multi-Agent Path Finding (MAPF) algorithm to search homotopy classes. 
It first searches for a coarse initial guess using a large search step, identifying a specific homotopy class. 
Subsequent decentralized Sequential Quadratic Programming (SQP) refinement processes this guess, resolving minor collisions efficiently. 
Experimental results demonstrate that CSDO outperforms existing MVTP algorithms in large-scale, high-density scenarios, achieving up to a 95\% success rate in 50 m $\times$ 50 m random scenarios around one second. Source codes are released in \textcolor{blue} {https://github.com/YangSVM/CSDOTrajectoryPlanning}.


\end{abstract}


\section{Introduction}

\IEEEPARstart{M}{ulti-Vehicle} Trajectory Planning (MVTP) seeks to generate a set of collision-free trajectories for multiple vehicles, from current positions to pre-set goals in a known unstructured environment, while minimizing travel time \cite{li_optimal_2021}. It is a fundamental problem with diverse applications, such as cooperative parking and warehouse automation. In practical applications, there is a need to efficiently obtain solutions within a limited time \cite{huang2023general}. As a non-convex optimization problem, MVTP necessitates a trade-off between solution quality and computational efficiency \cite{wen_cl-mapf_2022}. Particularly in scenarios involving a large number of vehicles, the frequency of vehicle-to-obstacle and vehicle-to-vehicle conflicts increases, complicating the search for optimal or even feasible solutions \cite{li_efficient_2021}. This work aims to develop an efficient algorithm that quickly finds feasible solutions with a high success rate for large-scale MVTP problems.


\subsection{Related Work}

Existing works struggle to find feasible solutions quickly at large scales, as shown in Table \ref{tab:mvtp}. The main challenge lies in efficiently exploring various homotopy classes \cite{li_optimal_2021}.
The homotopy class can be loosely defined as a set of solutions that are capable of continuous deformation into one another, without intersecting obstacles or other agents \cite{park2015homotopy}. 
Different homotopy classes can be seem as combinations of various routes and agent behaviors.
The quality of local optimal solutions within each homotopy class varies significantly. In large-scale scenarios, only a few homotopy classes might contain feasible solutions, making the exploration of various homotopy classes essential  \cite{li_optimal_2021}. 
Therefore, we evaluate the scalability of current MVTP algorithms in dense space based on their ability to explore homotopy classes. 


\begin{table}[htpb]
\caption{MVTP Algorithms}
\label{tab:mvtp}
\centering
\resizebox{0.95\linewidth}{!}{
\begin{tabular}{llll}
\toprule
Methods              & Feature                                                                                  & Runtime       & Scalability    \\ 
\midrule
Coupled              & Optimal; Complete.                                                                       & Very slow     & Very small     \\
Distributed  & Lack of cooperation.                                                                     & Fast          & Small          \\
Sampling based       & \begin{tabular}[c]{@{}l@{}}Probabilistic complete;\\ Asymptotically optimal\end{tabular} & Relative fast & Relative large \\
Constraint reduction & Optimal.                                                                                 & Relative slow & Large          \\
Tube construction    & Highly rely on initial guess.                                                            & Fast          & Small          \\
Grid search based    & \begin{tabular}[c]{@{}l@{}}Underuse efficient MAPF;\\ Step size trade-off.\end{tabular}  & Variable      & Large          \\ \bottomrule
\end{tabular}
}
\end{table}

\textit{Coupled planning} methods \cite{li_centralized_2017} treat all vehicles as a single, high dimensional agent. This approach relies solely on the optimizer's capability to traverse between different homotopy classes. While coupled planning methods guarantee completeness and optimality, the computational complexity increases rapidly with the growing number of non-convex constraints. 
In general, coupled planning methods exhibit poor scalability.

\textit{Distributed Planning} methods address MVTP in a single-agent manner, treating others as moving obstacles \cite{ma_decentralized_2023, luis_online_2020, alonso-mora_cooperative_2018}, or achieve collision avoidance through communication \cite{ferrantiDistributedNonlinearTrajectory2023, reyFullyDecentralizedADMM2018}. They have high efficiency in sparse scenarios but struggle with coordination, restricting the exploration of homotopy classes. In practice, these methods often struggle to generate high-quality collaboration, and the success rate decreases as scale increases, particularly in obstacle-dense scenarios.
\IEEEpubidadjcol

\textit{Sampling-based methods} \cite{soloveyFindingNeedleExponential2016,lukyanenkoProbabilisticMotionPlanning2023,shomeDRRTScalableInformed2020} mainly extend the Probabilistic Road Map (PRM) and Rapidly-exploring Random Tree (RRT) to multi-robot systems. These methods can provide probabilistic completeness and even asymptotical optimality. However, in crowded scenarios, these methods require a large number of samples, which can still lead to timeouts.

\textit{Constraint reduction} \cite{li_optimal_2021,ouyang_fast_2022, chen_decoupled_2015} dynamically adjusts the problem's complexity by adding or removing constraints, continuously approaching a feasible or even optimal solution. 
They achieve transitions between different homotopy classes by solving different nonlinear programming problems (NLP). However, solving NLP can be time-consuming.

\textit{Tube construction} methods \cite{honig_trajectory_2018, park_efficient_2020,shi2021neural} construct a safe corridor for each vehicle so the vehicle can be separated from the obstacles and other vehicles. 
Tube construction's solution is strictly homotopic to the reference trajectories. Therefore, it only searches limited homotopy classes and performs poorly without an approximately feasible initial guess.

\textit{Grid search} \cite{li_efficient_2021, wen_cl-mapf_2022} based methods discretize vehicle poses, actions, and space, utilizing a search algorithm to find discrete trajectories. This search algorithm is closely linked to a well-studied problem known as Multi-Agent Path Finding (MAPF), focusing on planning collision-free paths for multiple agents in a grid-like environment while minimizing travel time. 
Despite the NP-hard nature of MAPF, various efficient sub-optimal algorithms can generate paths for hundreds of agents in under a second \cite{ma_searching_2019}, aligning with MVTP's need for finding feasible solutions efficiently. However, the potential of these efficient MAPF algorithms remains largely unexplored in the MVTP field \cite{okumura2022priority}. Moreover, akin to single-agent grid searching motion planning algorithms, when the search step size is too small, the search space becomes too large, posing challenges for real-time requirements. Conversely, when the search step size is too large, the solution space diminishes, making it challenging to find a solution, and collisions may occur between search steps, rendering the solution infeasible. Therefore, grid search-based methods are more suitable for generating a coarse initial guess containing homotopy class information than directly generating fine solutions.


\subsection{Motivations and Contributions}



Based on the aforementioned literature review, existing methods search limited homotopy classes or search them inefficiently.
In this letter, we propose using an efficient MAPF solver to explore various coarse initial guesses with a large step size, which implictly encode different homotopy classes. After obtaining an initial guess that contains a specific homotopy class, decentralized optimization is employed to quickly generate a nearby kinematically feasible solution, ultimately achieving rapid generation of feasible solutions within a limited time.


 Accordingly, the main contributions are outlined as follows.
\begin{enumerate}
\item CSDO, an efficient, scalable multi-vehicle trajectory planning algorithm, employs a hierarchical framework to enhance search capabilities across diverse homotopy classes. Experiments demonstrate CSDO outperforms existing methods in random scenarios, especially in large scale and high-density environments.

\item A seamless adaptation of the priority-based search method from the MAPF domain into the complex non-holonomic MVTP problems, enables efficient exploration for feasible or near-feasible solutions.

\item An efficient distributed local solver is introduced. Given a homotopically correct reference solution, the local solver can generate feasible solutions quickly.
\end{enumerate}

\section{Problem Definition}

The MVTP problem can be defined by a ten element tuple $\langle M,\mathcal{W},\mathcal{O},z,\mathcal{R},s,g,f,\mathcal{T}, X \rangle$. Consider a system consisting of $M$ front-steering agents {$a^{(1)}, a^{(2)}, ..., a^{(M)}$} operating in a continuous planar workspace $\mathcal{W} \subset \mathbb{R}^2$. For simplicity, we use $[M]$ to denote the set $\{1,2,...,M\}$ and superscript $(i)$ to represent the variable related to agent $a^{(i)}$. There are some random static obstacles lying in the environment and occupying the workspace $\mathcal{O}$. $z=[x, y, \theta, \phi]^T \in \mathbb{R}^4$ refers to the state, where $(x, y)$ is the position of rear axis center, $\theta$ is yaw angle and $\phi$ is front-wheel steering angle. 
The control input is denoted as $u=[v,\omega]^T \in \mathbb{R}^2$, where $\omega=\dot\phi$, $v$ is the velocity. Agent $a^{(i)}$'s trajectory is represented by a sequence of its states sampled at fixed time interval $\Delta t$. It is denoted as $\mathcal{T}^{(i)} = [z_0^{(i)}, z_1^{(i)}, ..., z_{\tau_f^{(i)}}^{(i)}]$, where the $\tau_f^{(i)}+1$ is the number of states in the trajectory. 
For one MVTP task, $\mathcal{T}^{(i)}$ need to start from the start state $s^{(i)}$ and end at the goal state $g^{(i)}$.
\begin{equation} \label{eq:boundary_ocp}
    z^{(i)}_0 = s^{(i)}, z^{(i)}_{\tau_f^{(i)}} = g^{(i)}, \forall i \in [M].
\end{equation}
The task finish time is $makespan$ $\tau_f$, where $\tau_f = \max_{i \in [M]} \tau_f^{(i)}$.
It is assumed that the agent waits at the goal until all the agents have reached their goals, i.e. $z_t^{(i)} = g^{(i)}, \forall \tau_f^{(i)} 
\leq t \leq \tau_f$.
The planned trajectory $\mathcal{T}$ should be kinematic feasible for the Ackermann-steering model $f$, i.e.,
\begin{equation}\label{eq:kine_ocp}
\begin{aligned}
z_{t+1} &= z_{t} + \begin{bmatrix}
    v_t \cos \theta_t \\
    v_t \sin \theta_t \\
    v_t \tan(\phi_t)/L \\
    \omega \\
 \end{bmatrix} \Delta t, 
 \forall 0 \leq t < \tau_f,
\end{aligned}
\end{equation}

\begin{equation} \label{eq:control_max_ocp}
 \left | v_t \right | \leq v_{max}, \left | \omega_t \right | \leq \omega_{max}, 
 \forall 0 \leq  t < \tau_f,
\end{equation}
\begin{equation} \label{eq:phi_max_ocp}
 \left | \phi_t \right | \leq \phi_{max},  
 \forall 0 \leq  t \leq \tau_f ,
\end{equation}
where the $L$ is the vehicle's wheelbase. 
We use agent occupancy function $\mathcal{R}(z) : \mathbb{R}^4 \rightarrow \mathcal{W}$ to represent the workspace occupied by the agent's body at state $z$. The agents cannot collide with static obstacles or any other agents, i.e.,
\begin{equation} \label{eq:static_ocp}
    \mathcal{R}(z^{(i)}_{t}) \cap \mathcal{O}  = \emptyset, \forall t \geq 0, \forall i\in [M],
\end{equation}
\begin{equation} \label{eq:inter_ocp}
    \mathcal{R}(z^{(i)}_{t}) \cap \mathcal{R}(z^{(j)}_{t}) = \emptyset, \forall t \geq 0, \forall i,j\in [M], i \neq j.
\end{equation}

The solution plan $X$ comprises collision-free trajectories and control inputs of all agents. The solution quality is measured by $\tau_f$. Considering the summarized elements, a traditional optimal control problem \cite{li_optimal_2021} can be formulated as
\begin{equation}\label{eq:ocp}
\begin{aligned}
\min_{X, \tau_f} & \quad \quad \tau_f\\
\textrm{s.t.}& \quad  \textrm{Boundary Constraints (\ref{eq:boundary_ocp})},\\
  & \quad \textrm{Kinematic Constraints (\ref{eq:kine_ocp}), (\ref{eq:control_max_ocp}), (\ref{eq:phi_max_ocp}),}  \\
  & \quad \textrm{Static Collision Constraints (\ref{eq:static_ocp})}, \\
  & \quad \textrm{Inter-Agent Collision Constraints (\ref{eq:inter_ocp})}. 
\end{aligned}
\end{equation}

\section{Method}

\begin{figure*}[htpb]
    \centering
    \includegraphics[width=0.9\linewidth]{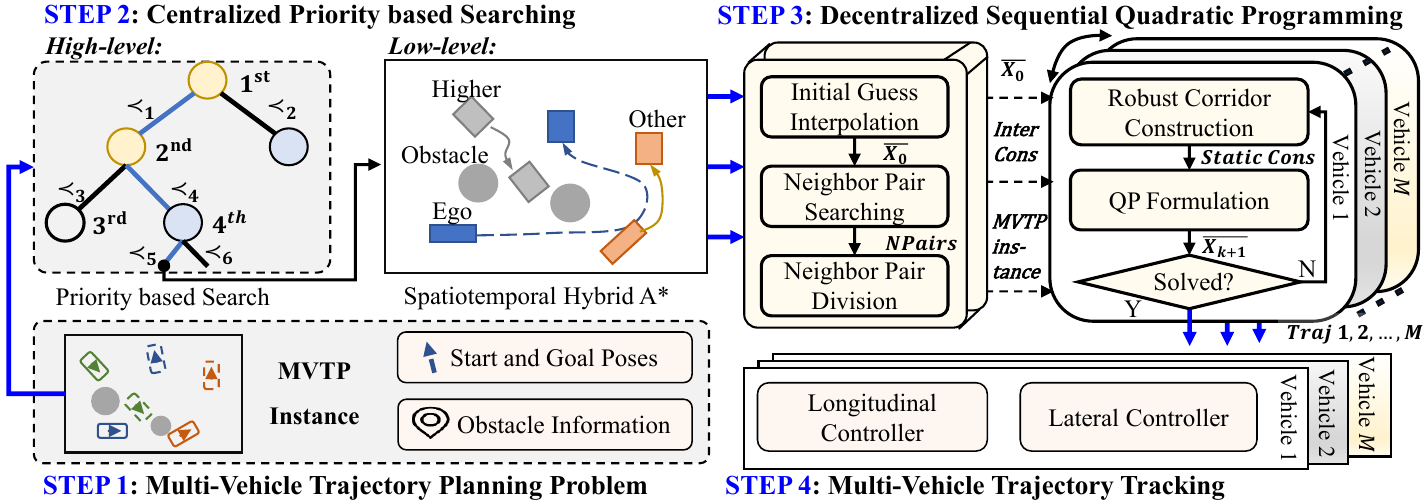} 
    \vspace{-3mm}
    \caption{The CSDO framework for multi-vehicle
trajectory planning.}
    \label{fig:Fig1}
    \vspace{-1mm}
\end{figure*}

The overall CSDO framework is illustrated in Fig. \ref{fig:Fig1}. Upon receiving the start poses, goal poses, and obstacle information, these components are combined to form a Multi-Vehicle Trajectory Planning (MVTP) instance. Subsequently, the centralized priority based searching phase generates coarse trajectories as an initial guess. The decentralized Sequential Quadratic Programming (SQP) refinement follows, where inter-vehicle constraints are decomposed and sent to multiple vehicles. Each vehicle utilizes the SQP solver to derive its trajectory and start to tracking simultaneously to reach their respective  goals.

\subsection{Centralized Priority based Searching}
\begin{figure}[htpb]
    \centering
    \includegraphics[width=1\linewidth]{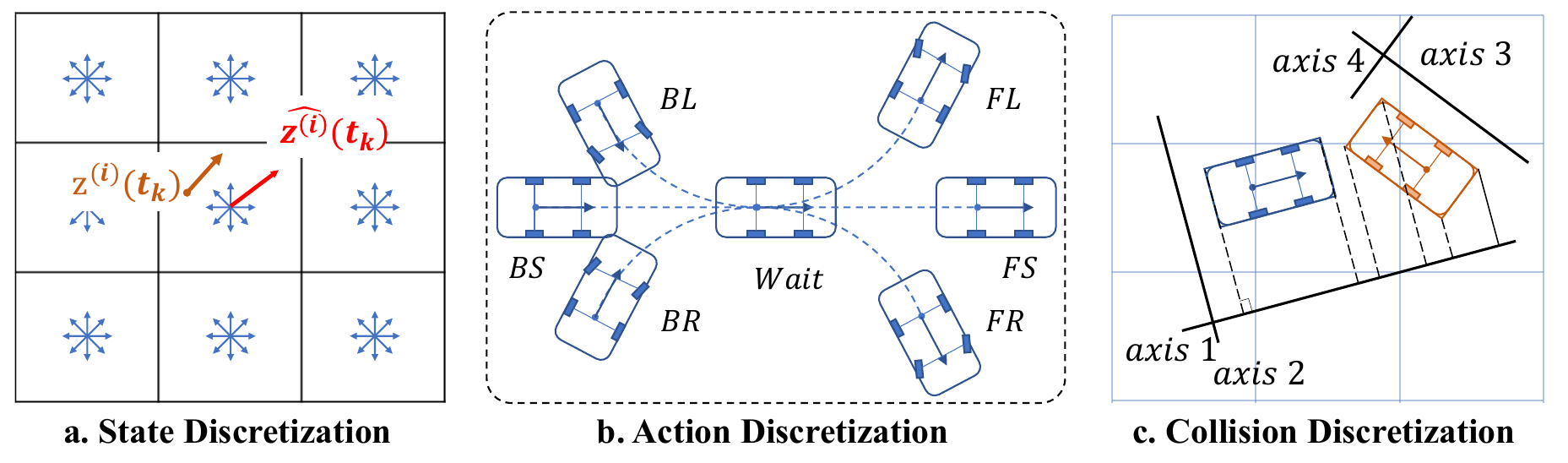} 
    \vspace{-3mm}
    \caption{MVTP discretization process.}
    \label{fig:Fig2}
    \vspace{-3mm}
\end{figure}


By discretizing the MVTP problem, we can utilize search algorithms to find an initial guess. As in Fig. \ref{fig:Fig2}, the discretizing process consists of state $z$ discretization, kinematic model $f$ discretization and agent collision detection implementation in formula (\ref{eq:static_ocp}-\ref{eq:inter_ocp}) as in single vehicle search algorithm hybrid A* \cite{dolgov_path_2010}. 
The discrete state $\hat z = [\hat x, \hat y, \hat \theta, 0]^T$, which contains the closest grid center position and discrete yaw angle, i.e., $\hat{z} = \argmin_{\hat z} || z - \hat z ||^2$. As in Fig. \ref{fig:Fig2}(a), $\hat z^{(i)}(t_k)$ is the discrete state of $z^{(i)}(t_k)$;
The kinematic model $f$ is simplified to an action set and limited to constant speed due to time complexity. $Actions=\{FL, FS, FR, BL, BS, BR, Wait\}$, which stand for front-max-steering-left, front-straight, front-max-steering-right, back-max-steering-left, back-straight, back-max-steering-right and wait respectively. Except for the $Wait$ action, all the actions, travel for the same step size $\Delta S$. In the grid search process, we adopt a large step size to search a coarse trajectory; Separating Axis Theorem is utilized the to check the collision at the sampling moment.


\begin{figure*}[htpb]
    \centering
    \includegraphics[width=0.9\linewidth]{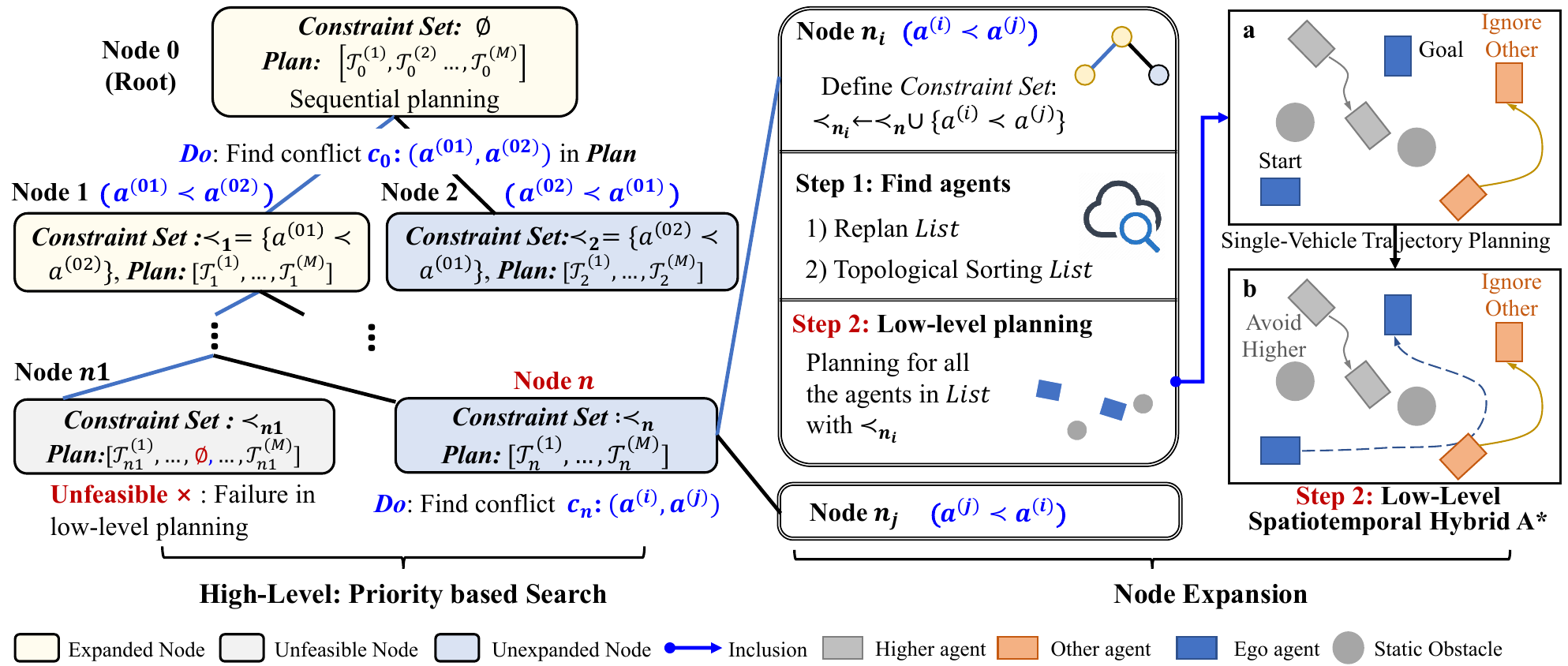} 
    \vspace{-3mm}
    \caption{Centralized priority based searching framework.}
    \label{fig:Fig3}
    \vspace{-1mm}
\end{figure*}

Our overall framework is illustrated in Fig. \ref{fig:Fig3}. It is worth noting that our CSDO architecture can achieve optimal or bounded sub-optimal solutions by applying corresponding MAPF algorithms. We choose PBS to find a feasible solution efficiently.
Centralized searching is divided into two layers of search. At the high level, each node represents a subproblem, and each node contains a \textit{constraint set} consisting of partial priority orders. Specifically, a partial priority order consists of two agents and refers to the avoidance relationship. For instance, if \( a^{(i)} \prec a^{(j)} \), then \( a^{(j)} \) has a lower priority than \( a^{(i)} \), and \( a^{(j)} \) treats \( a^{(i)} \) as a dynamic obstacle to avoid in the low-level planner. For any two agents, they may not have a relationship of \( a^{(i)} \prec a^{(j)} \) or \( a^{(j)} \prec a^{(i)} \), meaning they ignore each other and may have collisions. If collisions arise within a node's plan, two possible partial orders according to a specific conflict will be added to generate child nodes until collisions are entirely resolved.

\subsubsection{High-level search}

\begin{algorithm}[htpb]
    \caption{High level search}\label{algo:pbs}
        \SetKwInOut{Input}{Input}

        \Input{MAPF instance}
        $Root \gets $GENERATEROOT() \tcp*{$\prec_{Root}=\emptyset$}
        STACK $\gets \{Root\}$\;
        \While{$Stack$ $\neq \emptyset$}{
            $n \gets $ STACK.pop()\;
            \If{ $n.collisions$ = $\emptyset$}{
                    \Return $n.plan$;
            }
            $(a^{(i)}, a^{(j)}) \gets$ one colliding agent pair in $n.plan$\;
            \tcc{Node expansion.}
            $n_i, n_j \gets n$ \tcp*{Two copies of $n$} 
            \ForEach{ $(x, y) \in \{(i, j), (j, i)\}$}{
                $\pmb{\prec}_{n_x}\gets  \pmb{\prec}_{n} \cup \{a^{(x)} \prec a^{(y)} \}$ \;
                UPDATEPLAN($n_x$)\;
            }
            Insert \textbf{feasible} nodes into STACK in descending order of their makespan\;
        }
         
    \Return false;
    
\end{algorithm}
\vskip -1mm


We adapt PBS \cite{ma_searching_2019, li_intersection_2023} to address our problem as follows. As in algorithm \ref{algo:pbs}, the root node [Line 1] is initialized with an empty set of priority orders, but we employ a \textbf{warm start} technique to speed up the search process. Within the root node, we attempt to plan the agents sequentially, treating the previously planned agents as dynamic obstacles. If any agent encounters planning failure due to obstruction by preceding agents, we allow it to plan its trajectory freely;
When expanding a node, we check for collisions between each vehicle pair. If no collisions are detected, the node is considered the final result [Lines 5 to 6]. Otherwise, we select a pair of colliding vehicles, $a^{(i)}$ and $a^{(j)}$ [Line 7].
The following part describes the detailed procedure of node expansion as shown in Fig. \ref{fig:Fig3}. Two constraints, $a^{(i)} \prec a^{(j)}$ and $a^{(j)} \prec a^{(i)}$, are created and added separately to form the new child nodes' constraint sets. 
For example, $\prec_{n_i} = \prec_{n} \cup \{a^{(i)} \prec a^{(j)} \}$ as in node $n_i$ [Lines 10 to 11]. One straightforward replanning method involves replanning all the agents according to $n_x$. However, to update the plan [Line 12] without redundant replanning, we first identify the agents violating the new constraint set $\prec_{n_x}$ and perform a topological sorting on the agents. Next, we replan the agents from higher to lower priority using the low-level planner. Finally, feasible child nodes are inserted [Line 13] in the non-decreasing order of the planned makespan of the nodes.


\subsubsection{Low-level planner}
We directly adopt the complete and optimal spatiotemporal Hybrid A* (STHA*) from \cite{wen_cl-mapf_2022} as our low-level planner. Compared with Hybrid A*, STHA* adds a time dimension to deal with the dynamic obstacles.  Given a workspace $\mathcal{W}$ and static obstacle occupancy workspace $\mathcal{O}$, the higher priority agents' trajectories $highTrajs$, a predefined start state $s_i$ and goal state $g_i$, STHA*  will search the fastest trajectory when the solution space is not empty.

\subsubsection{Completeness and Optimality Analysis}

\begin{figure}[htpb]
    \centering
p    \includegraphics[width=0.95\linewidth]{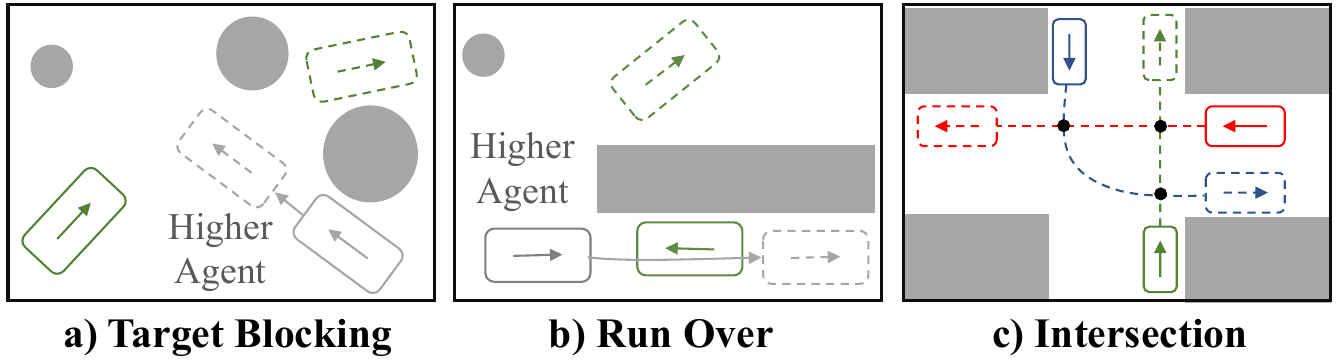} 
    \vspace{-3mm}
    \caption{PBS failed reason and well-formed scenarios.}
    \label{fig:pbs_complete}
\end{figure}

Prioritized planning can fail to find a solution due to inappropriate priority orders \cite{yangAttentionbasedPriorityLearning2024}. The only issues are target blocking and run-over, as illustrated in Fig. \ref{fig:pbs_complete}. Target blocking occurs when a high-priority agent reaches its goal early and blocks a low-priority agent between its current position and goal. Run-over happens when a low-priority agent has no possible trajectory to avoid a high-priority agent.
By utilizing tree search, PBS can explore all possible priority orders, making it \textbf{P-complete}. This capability allows PBS to greatly mitigate the above issues.
Furthermore, in well-formed problems, all priority orders can lead to a feasible solution. So PBS can quickly find a feasible and even near-optimal solution.
The key feature of well-formed problems is that agents can wait at their start and goal positions indefinitely without blocking other agents \cite{ma_searching_2019}.
In practice, well-formed problems are very common, such as in intersection coordination\cite{li_intersection_2023}. 
Additionally, by changing the search strategy from depth-first search to best-first search, PBS can find a \textbf{P-optimal} solution, meaning it can find the best solution quality among the priorities, being optimal or near-optimal in practice \cite{ma_searching_2019}.

\subsection{Decentralized SQP}

As shown in Fig. \ref{fig:Fig1}, after inputting the initial guess, the separation planes are constructed to serve as inter-vehicle constraints. Then, multiple distributed SQP processes are employed to generate all the trajectories.

\textbf{Notation}. For clarity, the bar symbol represents the constant value. The subscript 0 denotes the constant associated with initial guess. For instance, $\bar \theta^{(i)}_{t,0}$ is the yaw angle of agent $a^{(i)}$ at timestamp $t$ in the interpolated initial guess. 
When there is no ambiguity, we omit the corresponding superscript $(i)$ or subscript $t$ when referring to all agents or any timestamp. 

\subsubsection{Initial Guess Interpolation}
\textbf{Input} initial guess $\bar X_{raw0}$. \textbf{Output} interpolated initial guess $\bar X_0$.

\begin{figure}[htpb]
    \centering
    \includegraphics[width=1\linewidth]{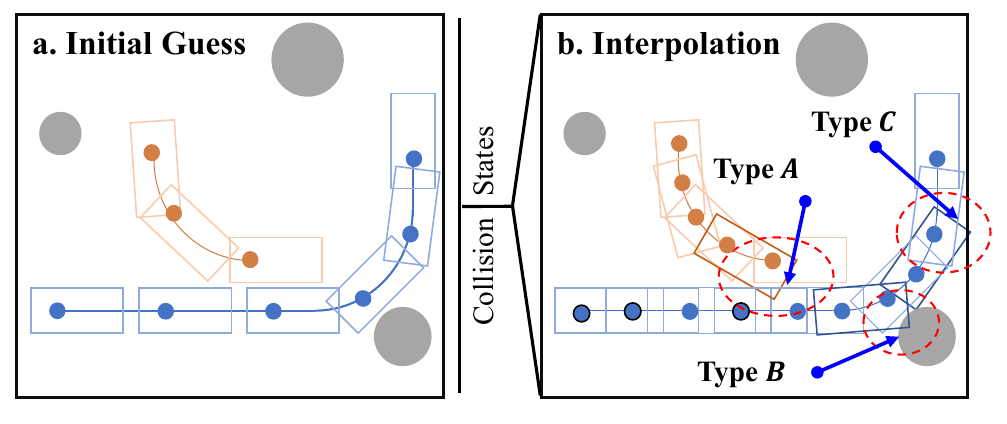} 
    \vspace{-5mm}
    \caption{The initial guess and collision states after interpolation.}
    \label{fig:Fig4}
    \vspace{-1mm}
\end{figure}

We interpolate each curve segment by inserting \( n_{interp} \) points into each segment. So the time interval $\Delta t$ of adjacent points can be calculated as 
$    \Delta t = \Delta S /((n_{interp}+1) v_{max})$.
Afterwards, the initial guess may encounter three types of minor collisions, as illustrated in Fig. \ref{fig:Fig4}: type A: collisions between two vehicles, type B: collisions with obstacles, and type C: off-map states.




\subsubsection{Neighbor Pair Searching}
\textbf{Input} Interpolated initial guess $\bar X_0$. \textbf{Output} Neighbor pairs $NPairs$.

\begin{figure}[htpb]
    \centering
    \includegraphics[width=1\linewidth]
    {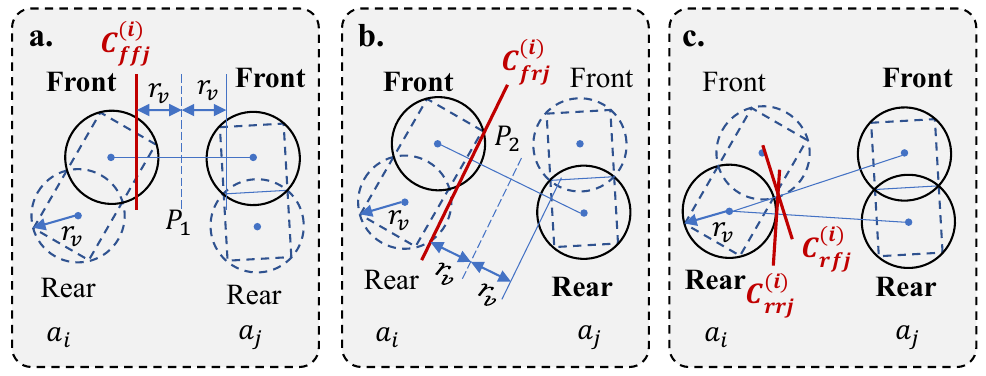} 
    \vspace{-5mm}
    \caption{Generate separating planes for neighbor pairs.}
    \label{fig:Fig5}
    \vspace{-2mm}
\end{figure}

To facilitate distance measuring for neighbor pair search and collision avoidance, we employ two uniformly distributed circles to cover the rectangular shape of the vehicle \cite{ouyang_fast_2022}. As illustrated in Fig. \ref{fig:Fig5}, the circle centers are positioned at the quadrant points. The formulas for calculating the centers and radii of the two circles are as follows.

\begin{subequations} \label{eq:state2disc}
\begin{equation}
x^{F} = x + L_{f2x} \cos{\theta} ,
y^{F} = y + L_{f2x}  \sin{\theta} ,
\end{equation}
\begin{equation}
x^{R} = x + L_{r2x} \cos{\theta},
y^{R} = y + L_{r2x}\sin{\theta},
\end{equation}
\begin{equation}
r_v = \frac{1}{2} \sqrt{\frac{1}{4}(L_F^2 + L_B^2) + L_B^2},
\end{equation}
\end{subequations}
where the $L_{f2x} = (3L_F - L_B)/4$ and $L_{r2x} = (L_F - 3L_B)/4$ are distance from the rear-axis center to the front and rear disc center, 
$L_F$ and $L_R$ are the distance from rear axis to the front bumper and rear bumper.
$Y^{F} = [x^{F}, y^{F}]^T$ and $Y^{R}=[x^{R}, y^{R}]^T$ are the center of the front and rear circle. $r_v$ is the radius of the circle. We denote $Y=[x^F,y^F,x^R,y^R]^T$ as the vector of the circles positions. 

Given a distance threshold $R_{trust}$, we iterate through the plan to search pairs of agents $a^{(i)}$ and $a^{(j)}$ with a distance less than $2\sqrt{2}R_{trust}$ at the same timestamp. The distance function between two states $(z^{(i)}_t, z^{(j)}_t)$ is defined as follows:
\begin{equation}
\begin{aligned}
    dist&(z^{(i)}_t, z^{(j)}_t) = \min( 
    \left \| Y^{F(i)}_t - Y^{F(j)}_t  \right \|, 
    \left \| Y^{F(i)}_t - Y^{R(j)}_t  \right \|, \\
  &\quad \left \| Y^{R(i)}_t - Y^{F(j)}_t  \right \|, 
    \left \| Y^{R(i)}_t - Y^{R(j)}_t  \right \| ) - 2r_v,
\end{aligned}
\end{equation}
where the $Y^{F(i)}_t$, $Y^{R(i)}_t$, $Y^{F(j)}_t$, $Y^{R(j)}_t$ are the agent $a^{(i)}$'s front disc center, back disc center, agent $a^{(j)}$'s front disc center and rear disc center respectively. The search result, denoted as $NPairs$, comprises neighbor pairs and their corresponding timestamps. $NPairs=\{(a^{(i)},a^{(j)},t)|dist(z^{(i)}_t, z^{(j)}_t) \leq 2\sqrt{2}R_{trust}, \forall i,j \in [M], i \neq j, \forall 0 \leq t \leq \tau_f \}$.

\subsubsection{Neighbor Pair Division}
\textbf{Input} neighbor pairs $NPairs$. \textbf{Output} inter-collision avoidance constraints (\ref{eq:inter} - \ref{eq:trust}).

After obtaining the neighbor pairs $NPairs$, we utilize the plane derived by the perpendicular bisector of the disc's center as the constraint for mutual avoidance between vehicles. Each neighbor pair $(a^{(i)},a^{(j)},t)$ generates 4 separation planes \{$C_{ffj}^{(i)}, C_{frj}^{(i)}, C_{rfj}^{(i)}, C_{rrj}^{(i)}$\} for agent $a^{(i)}$. Fig. \ref{fig:Fig5} illustrates the process. 
Calculating perpendicular bisectors, offset by a distance of $r_v$ to obtain the separation half-plane $C_{ffj}^{(i)}$. 
Similarly, we generate the front-to-rear, rear-to-front and rear-to-rear separation half-planes $C_{frj}^{(i)}$, $C_{rfj}^{(i)}$ and $C_{rrj}^{(i)}$, respectively.

\begin{equation}\label{eq:inter}
   \bar A_{c,0} Y \leq \overrightarrow{0},
\end{equation}
where the $\bar A_{c,0}$ denotes the corresponding half planes. Notably, this method seamlessly adapts RSFC \cite{parkOnlineTrajectoryPlanning2021} under the assumption of two-circle approximation and discrete-time collision detection.

To focus the search within the neighbor of the initial guess, we restrict the variation range of the disk to $R_{trust}$, i.e.,
\begin{equation}\label{eq:trust}
    \left | Y - \bar Y_{0} \right |\leq R_{trust}\overrightarrow{1}.
\end{equation}

\begin{remark}[]
Constraints (\ref{eq:inter})-(\ref{eq:trust}) are equivalent to constraint (\ref{eq:inter_ocp}), guaranteeing that there are no collisions between vehicles at each discrete time step.
\end{remark}
\begin{proof}
For neighbor pairs, they are separated by the planes as in constraint (\ref{eq:inter}). For non-neighbor pairs, they are separated by the variation range constraint (\ref{eq:trust}).
\end{proof}
\begin{remark}[]
Constraints (\ref{eq:inter})-(\ref{eq:trust}) decouple the trajectory variables between different vehicles.
\end{remark}
\begin{proof}
$\bar A_{c,0}$ and $\bar Y_{0}$ are constants determined by the initial guess. For any two different vehicles $a^{(i)}$ and $a^{(j)}$, their variables $z^{(i)}$ and $z^{(j)}$ will not appear in the same inequality.
\end{proof}

The above process decouples agents for distributed problem-solving. Without loss of generality, we describe the processing procedure for agent $a^{(i)}$ in the following steps.
This process is executed repeatedly until the stop criteria are met.
\textbf{Notation}. The subscript $k$ represents the k-th iteration, with $k=0$ indicating the initial guess.


\subsubsection{Robust Corridor Construction}

\textbf{Input} obstacles $\mathcal{O}$ and $\bar X^{(i)}_{k}$. \textbf{Output} static collision avoidance constraints (\ref{eq:static}).

To handle the static obstacle avoidance constraint (\ref{eq:static_ocp}), we adapt the method from \cite{li_optimization-based_2022} to generate a corridor along the last iteration solution $\bar X_k^{(i)}$ of agent $a^{(i)}$. 

\begin{figure}[htpb]
    \centering
    \includegraphics[width=0.95\linewidth]{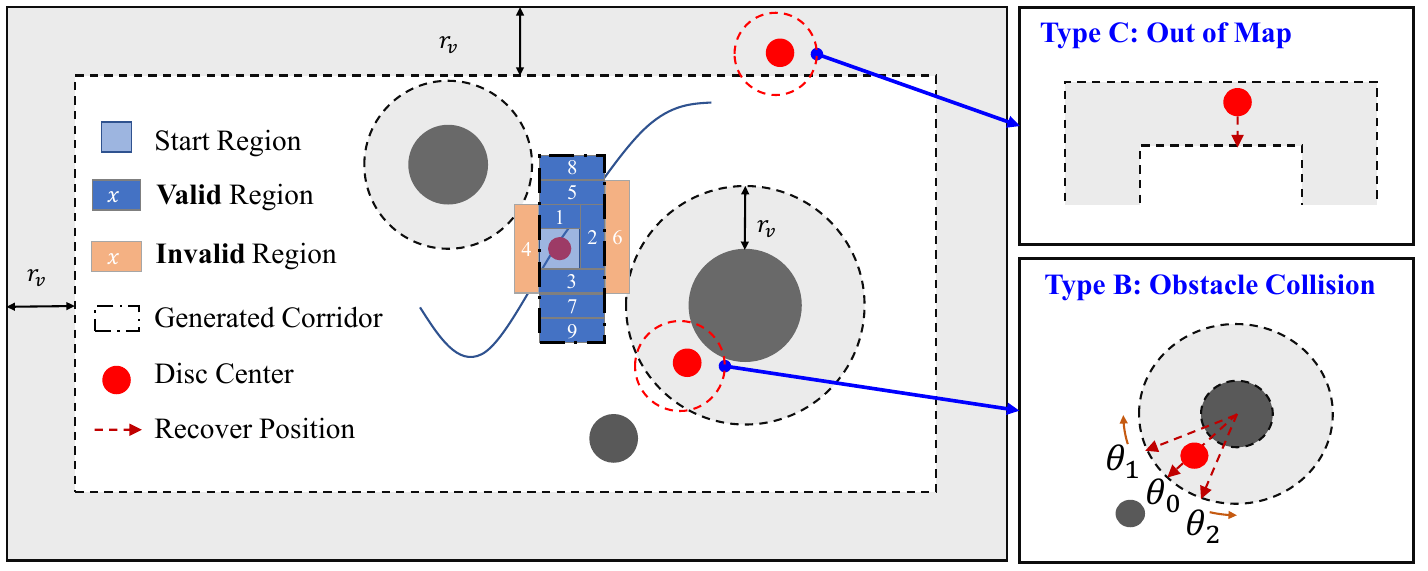} 
    \vspace{-3mm}
    \caption{Corridor construction from a legal start point.}
    \label{fig:corridorConstruction}
    \vspace{-1mm}
\end{figure}

For clarity, ensuring that a disc with radii $r_v$ does not go out of the map is equivalent to maintaining a distance of $r_v$ from the border. As in Fig. \ref{fig:corridorConstruction}, we erode the map by a distance of $r_v$ to define the safety space enclosed by the dotted line. Similarly, we dilate the obstacles by a distance of $r_v$.

As illustrated in Fig. \ref{fig:corridorConstruction}, we sequentially extend the empty box clockwise in all four directions until it encounters dilated obstacles, the eroded map boundary, or reaches the maximum allowed length. Details can be found in \cite{li_optimization-based_2022}.

Note that our initial point may be in an out-of-map or colliding state, as previously mentioned in Fig. \ref{fig:Fig4}, causing the algorithm to immediately return an empty box. Therefore, we must relocate the initial point to a safe position before generating the box. For initial points that are out-of-map, we project them onto the map boundary. If the original or projected state collides with obstacles, we move the point outside of the grey circle. If it remains unsafe due to collisions with other obstacles, we gradually rotate it around the obstacle center until it becomes safe. The corridor constraints can be summarized as follows,
\begin{equation} \label{eq:static}
   \bar Y^{(i)}_{min,t,k} \leq \bar A_{static,t,k}^{(i)}Y^{(i)}_{t,k+1} \leq \bar Y^{(i)}_{max,t,k}, \forall 0 \leq t \leq \tau_f,
\end{equation}
where $\bar A_{static,t,k}^{(i)}$ denotes the generated corridor as in Fig. \ref{fig:corridorConstruction}.

\subsubsection{QP Formulation}
\textbf{Input} constraints (\ref{eq:inter}-\ref{eq:static}) and last solution $\bar X^{(i)}_{k}$. \textbf{Output} Refine agent solution $\bar X^{(i)}_{k + 1}$.

Finally, we linearize the kinematic constraints. This enables the smooth speed profile generation. The linearization error will be alleviated by the following sequential refinements. The objective function is set to minimize changes in velocity and steering wheel angle, aiming to smooth the trajectory.
The objective function is as follows,
\begin{equation}\label{eq:cost}
    J = \Sigma_{t} (\alpha_v (\Delta v_{t,k+1}^{(i)})^2 + \alpha_{\omega} (\omega_{t,k+1}^{(i)} )^2), 
\end{equation}
where $\alpha_v$ and $\alpha_\omega$ are the weighting parameters.

The kinematic constraints are linearized as follows:
\begin{subequations}\label{eq:linear_kine}
\begin{equation}
z_{t+1,k+1}^{(i)} = \bar A_{t,k}^{(i)} z_{t,k+1}^{(i)} +\bar B_{t,k}^{(i)} u_{t,k+1}^{(i)} + \bar c_{t,k}^{(i)}, \forall 0 \leq t < \tau_f
\end{equation}
\begin{equation}
    \bar A_{t,k}^{(i)} = \begin{bmatrix}
     1& 0 & -\bar v_{t,k}^{(i)} \sin \bar \theta_{t,k}^{(i)} *\Delta t & 0 \\ 
     0& 1 & -\bar v_{t,k}^{(i)} \cos \bar \theta_{t,k}^{(i)} *\Delta t & 0\\ 
     0& 0 & 1 & \frac{\bar v_{t,k}^{(i)}*\Delta t}{L \cos^2 \bar \phi_{t,k}^{(i)}}  \\ 
     0& 0 & 0 & 1
\end{bmatrix} ,
\end{equation}
\begin{equation}
\bar B_{t,k}^{(i)} = \begin{bmatrix}
 \cos \bar \theta_{t,k}^{(i)} \Delta t& \sin\bar \theta_{t,k}^{(i)} \Delta t &  \frac{\tan \bar \phi_{t,k}^{(i)} \Delta t}{L} &0 \\ 
 0& 0&  0&  \Delta t 
\end{bmatrix}^T,
\end{equation}
\begin{equation}    
\bar c_{t,k}^{(i)} = [\bar \theta_{t,k}^{(i)} \bar v_{t,k}^{(i)} \sin \hat{\theta_{t,k}^{(i)}} \Delta t, -\bar \theta_{t,k}^{(i)} \bar v_{t,k}^{(i)} \cos \hat{\theta_{t,k}^{(i)}} \Delta t, -\frac{\bar \phi_{t,k}^{(i)} \Delta t}{L \cos^2{\phi_{t,k}^{(i)}}}]^T, 
\label{eq:end}
\end{equation}
\end{subequations}
where $(\bar A_{t,k}^{(i)}, \bar B_{t,k}^{(i)}, \bar c_{t,k}^{(i)})$ are the associated coefficients.

To handle the non-linear calculation from state $z$ disc center positions $Y$, we need to linearize the Eq. (\ref{eq:state2disc}), i.e.,
\begin{subequations}\label{eq:linearState2Y}
\begin{equation}
    Y_{t,k+1}^{(i)} = \bar D_{t,k}^{(i)} z_{t,k+1}^{(i)} + \bar e_{t,k}^{(i)}, \forall 0 \leq t \leq \tau_f
\end{equation}
\begin{equation}
    \bar D_{t,k}^{(i)} = \begin{bmatrix}
    1 & 0 & -L_{f2x} \sin \bar \theta_{t,k}^{(i)} & 0 \\
    0 & 1 & L_{f2x} \cos \bar \theta_{t,k}^{(i)} & 0 \\
    1 & 0 & -L_{r2x} \sin \bar \theta_{t,k}^{(i)} & 0 \\
    0 & 1 & L_{r2x} \cos \bar \theta_{t,k}^{(i)} & 0      \\
    \end{bmatrix}, \\
\end{equation}
\begin{equation}
    \bar e_{t,k}^{(i)} = \begin{bmatrix}
    L_{f2x}(\cos{\bar \theta_{t,k}^{(i)}} + \bar \theta_{t,k}^{(i)} \sin \bar \theta_{t,k}^{(i)})  \\
    L_{f2x}(\sin{\bar \theta_{t,k}^{(i)}} - \bar \theta_{t,k}^{(i)} \cos \bar \theta_{t,k}^{(i)})  \\
    L_{r2x}(\cos{\bar \theta_{t,k}^{(i)}} + \bar \theta_{t,k}^{(i)} \sin \bar \theta_{t,k}^{(i)})  \\
    L_{r2x}(\sin{\bar \theta_{t,k}^{(i)}} - \bar \theta_{t,k}^{(i)} \cos \bar \theta_{t,k}^{(i)})  
\end{bmatrix}. \\
\end{equation}
\end{subequations}

With the aforementioned elements summarized, a complete QP is formulated as follows,

\begin{equation}
\begin{aligned}
\min_{{X}^{(i)}} \quad &  J \\
\textrm{s.t.} \quad & z_{0,k+1}^{(i)}=[s^{(i)T}, 0]^T, z_{\tau_f,k+1}^{(i)}=[g^{(i)T}, 0]^T, \\
  & \left | v^{(i)}_{t,k+1} \right | \leq v_{max}, \left | \omega^{(i)}_{t,k+1} \right | \leq \omega_{max}, \forall 0 \leq  t < \tau_f,\\
  & \left | \phi^{(i)}_{t,k+1} \right | \leq \phi_{max},  \forall 0 \leq  t \leq \tau_f,\\
  & \textrm{Constraints \quad (\ref{eq:inter})-(\ref{eq:linearState2Y})} \quad \textrm{related to $a^{(i)}$}
\end{aligned}
\end{equation}

\subsubsection{Stop Criteria}
When the $|| \bar X_{k+1} - \bar X_k ||_2$ is less than the given threshold, or the plan is feasible, we stop the iteration.

\section{Experiment}
To demonstrate the effectiveness of our method, we conduct experiments on randomly generated obstructed maps as well as obstacle-free maps. We incrementally increased the number of agents in the problem, resulting in more congested maps, with a specific emphasis on showcasing the effectiveness of our approach in addressing large-scale MVTP problems.

\subsection{Simulation Settings} 
The benchmark consists of various map sets with different sizes (50 m , 100 m), varying numbers of agents (5 to 100), and includes both random and room-like maps. Each map set contains 60 instances, resulting in a total of 2100 test instances in this testing benchmark.
We assume all agents are homogeneous and share the following parameters: the vehicle's shape is 3 m × 2 m, 
and  it has a maximum speed of 1 m/s.
All algorithms are assessed using their respective open-source implementations, with most executed on an Intel Xeon Gold 622R CPU at 2.90 GHz in C++, while Fast-ASCO runs on an Intel Core i7-9750H CPU at 2.6 GHz in Matlab.

\subsection{Simulation Results}
\begin{figure*}[htpb]
 \begin{center}        
 \begin{tabular}{c}
      \includegraphics[width=0.9\linewidth]{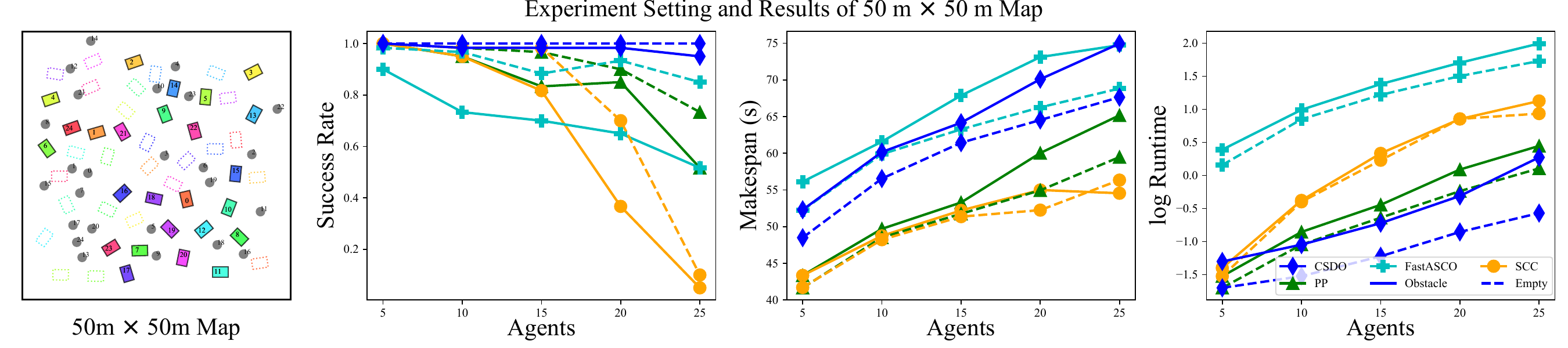} \\ 
      \includegraphics[width=0.9\linewidth]{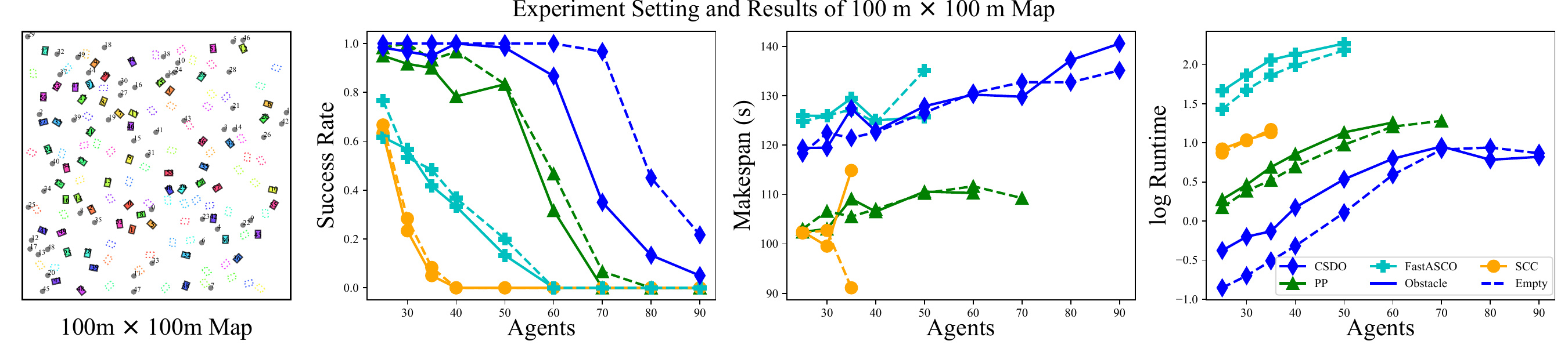}
 \end{tabular}
 \end{center}

 \vspace{-3mm}
 \caption{Simulation results on random maps. The solid line represents obstructed scenarios, and the dotted line represents obstacle-free scenarios.}
 \label{fig:map100}
 \vspace{-1mm}
\end{figure*}

\begin{table*}[htpb]
\caption{Simulation results on random room-like 100 m $\times$ 100 m map.} \label{tab:result}
\vspace{-2mm}
\centering
\resizebox{\linewidth}{!}{
\begin{tabular}{|l|ccccc|ccccc|ccccc|}
\hline
Method    & \multicolumn{5}{c|}{Success Rate $\uparrow$}                                                  & \multicolumn{5}{c|}{Runtime (s) $\downarrow$}                                 & \multicolumn{5}{c|}{Makespan (s) $\downarrow$}                                          \\ \hline
\# Agents & 10                & 20               & 30               & 40               & 50               & 10            & 20            & 30            & 40            & 50            & 10              & 20              & 30              & 40              & 50              \\ \hline
SCC       & 93.33\%           & 33.33\%          & 0                & 0                & 0                & 4.43          & 8.10          & -             & -             & -             & \textbf{133.02} & \textbf{115.01} & \textbf{-}      & \textbf{-}      & \textbf{-}      \\
PP        & \textbf{100.00\%} & 91.67\%          & 61.67\%          & 18.33\%          & 6.67\%           & 2.11          & 7.34          & 11.02         & 13.19         & 13.53         & 135.41          & 142.00          & 131.73          & 127.35          & \textbf{111.80} \\
FastASCO  & 30.00\%           & 25\%             & 25\%             & 5\%              & 5\%              & 7.42          & 35.05         & 89.28         & 154.61        & 378.39        & \textbf{93.65}  & \textbf{111.3}  & \textbf{110.83} & \textbf{111.64} & 112.94          \\
CSDO      & \textbf{100.00\%} & \textbf{98.33\%} & \textbf{80.00\%} & \textbf{60.00\%} & \textbf{33.33\%} & \textbf{1.19} & \textbf{3.64} & \textbf{5.49} & \textbf{7.86} & \textbf{9.19} & 171.78          & 187.37          & 177.71          & 181.57          & 178.53          \\ \hline
\end{tabular}
}

\end{table*}

The performance of CSDO is evaluated through a comparative analysis with various MVTP algorithms, focusing on success rate, runtime, and solution quality (makespan). A general time limit of 20 s is applied, except for Fast-ASCO, which is allowed 200 s due to Matlab implementation. Figure \ref{fig:map100} presents the results for both map types, while Table \ref{tab:result} displays results for a map size of 50 m $\times$ 50 m.

Seq-CL-CBS (SCC) \cite{wen_cl-mapf_2022} is a grid search-based method and a prioritized planning version of the optimal MVTP method, CL-CBS, for large-scale problems. 
CL-CBS, leveraging the optimal MAPF algorithm Conflict based Search (CBS), forms the basis for  SCC, which organizes agents into several groups for sequential planning with CL-CBS; each subsequent group views the prior as dynamic obstacles. 

Prioritized Planning (PP) is a grid search-based prioritized planning method. It randomly assigns each agent an order and plans their trajectories sequentially based on this order. 
In theory, PP is not a complete or optimal method and may perform poorly if an inappropriate order is chosen. 


Fast ASCO \cite{ouyang_fast_2022}, an advanced ASCO variant \cite{li_optimal_2021}, excels in constraint reduction, offering an optimal solution despite the high runtime. In addition, it optimizes for both travel time and comfort, and has a detailed kinematics model. 

In our simulation, CSDO achieves the best success rate and runtime in general, whether on random maps or room-like maps, benefiting from PBS's efficiency and its hierarchical framework. 
Though SCC enhances scalability through sequential planning, its computation time still increases exponentially with large-scale problems due to CBS; 
PP scales better than SCC, and the solution quality does not deteriorate significantly. However, compared to CSDO, PP exhibits poorer performance in terms of success rate and runtime. CSDO relies on PBS, which searches various partial order sets and has higher completeness. 
Regarding solution quality, PP may outperform CSDO, as CSDO prioritizes success rate over makespan optimization;
As an optimal algorithm, Fast ASCO achieves superior success rates in large scale, outperforming near-optimal methods like SCC for groups of 20 and 25 agents in a 50 m square map. It achieves a longer makespan due to the optimization for comfort and the detailed kinematics model.

\subsection{Ablation Study and Limitation Analysis}
\vspace{-3mm}
\begin{table}[htpb]
\begin{adjustbox}{max width=0.47\textwidth, center}

\begin{threeparttable}

\caption{Ablation Study on 100m $\times$ 100m random map}
\label{tab:scale}
\centering

\begin{tabular}{lllll}
\toprule
\# Agents               & 30      & 50      & 70      & 90     \\ 
\midrule
SR$^{a}$: CSDO                & 96.67\% & 98.33\% & 35.00\% & 5.00\% \\
SR: CSDO w/o$^{d}$ DO$^{e}$         & 0       & 0       & 0       & 0      \\
SR: CSDO w/o warm start & 93.33\% & 95.00\% & 11.67\% & 0      \\
FR$^{b}$: DO failure rate$^{f}$     & 1.66\%  & 0       & 1.66\%  & 0      \\ 
\midrule
RT$^{c}$: CSDO (s)               & 0.63    & 3.42    & 8.99    & 6.62   \\
RT: CS w/o DO (s)                  & 0.49    & 3.24    & 8.74    & 6.43   \\
RT: CSDO w/o warm start (s) & 1.53    & 9.78    & 14.72   & -      \\ 
\bottomrule
\end{tabular}

\smallskip
\scriptsize
\begin{tablenotes}[para,flushleft,small]
\RaggedRight
\item[a] SR: Success Rate; 
\item[b] FR: Failure Rate;
\item[c] RT: Runtime;
\item[d] w/o: without;
\item[e]  CSDO w/o DO: Only use centralized search. The success rate  means the initial guess has no collision and can be seem as a feasible solution;
\item[f] DO failure rate: the percentage of cases where decentralized optimization fails to find one feasible solution despite a successful initial guess from centralized searching;
\end{tablenotes}
\end{threeparttable}
\end{adjustbox}
\vspace{-3mm}
\end{table}

Ablation experiments are conducted to validate the effectiveness of Decentralized Optimization (DO) and a warm start. Without DO, nearly all initial guesses exhibit minor collisions and are deemed infeasible. The runtime of centralized searching dominates within CSDO. Furthermore, the DO failure rate is less than 2\%, indicating that the primary completeness loss in CSDO is attributed to PBS. Thus, the necessity, speed, and effectiveness of DO are validated. The warm start technique contributes to varying degrees of improvement in both success rate and runtime metrics.

In summary, regarding limitations, the PBS can be P-complete and can be modified to be P-optimal. For the DO, the completeness drop is less than 2\% in simulations.

\subsection{Experimental Setup and Results}


\begin{figure}[htpb]
    \centering
   \begin{tabular}{cc}
      \adjustbox{valign=t}{\includegraphics[width=0.35\linewidth]{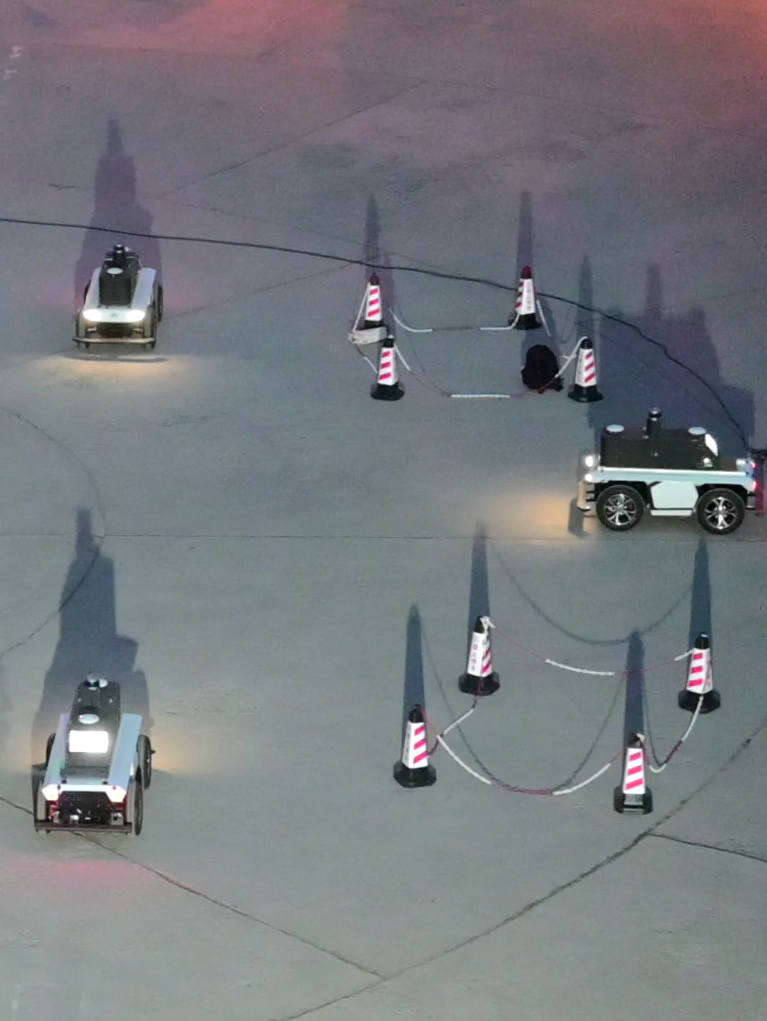}} &  
      \adjustbox{valign=t}{\includegraphics[width=0.55\linewidth]{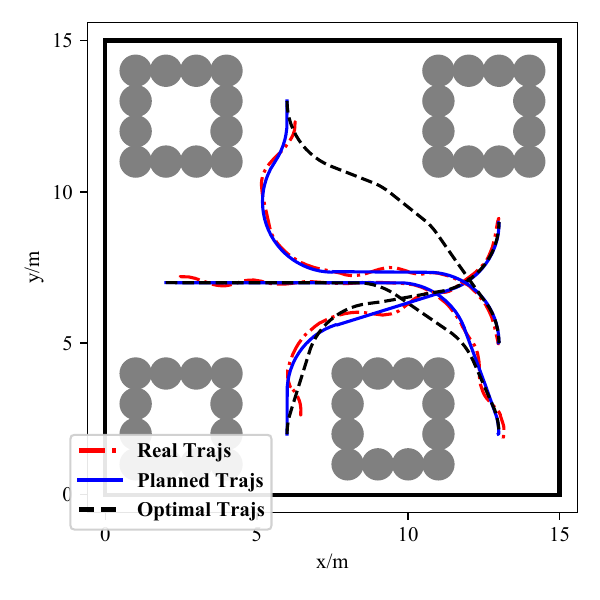}} \\
         (a)  & (b) 
    \end{tabular}
         \caption{Real world experiments and results on 15 m square map. (a) Experiment platform. (b) The vehicle real trajectories, CSDO planned trajectories and the optimal trajectories planned by CL-CBS on 15 m map.}
    \label{fig:fieldtest}
    \vspace{-3mm}
\end{figure}

\begin{table}[ht]

\caption{ Real-world experiment results}
\vspace{-3mm}
\label{tab:field}
\centering
\begin{tabular}{ccccc}
\toprule
Method & \multicolumn{2}{c}{15 m $\times$ 15 m map}          & \multicolumn{2}{c}{20 m $\times$ 20 m map}           \\ \cmidrule{2-5} 
                        & $\tau_f$ (s)$\downarrow$& Runtime (s)$\downarrow$ & $\tau_f$ (s)$\downarrow$ & Runtime (s)$\downarrow$ \\ \midrule
CL-CBS                  & \textbf{17.4 }                     & 10.075                    & \textbf{15.5}                      & 0.332                     \\
CSDO                    & 20.3                      & \textbf{0.623}                     & 16.8                      & \textbf{0.014}                    \\ \bottomrule
\end{tabular}
\end{table}

Experiments are conducted with 3 forward-only 1.9 m $\times$ 1.3 m vehicles in 15 m and 20 m square areas. The vehicles are positioned using differential GNSS. The map is pre-set and known. The trajectories are calculated on a typical laptop and then transmitted to the agents via WiFi. The trajectory tracking controller operates at 10 Hz, utilizing longitudinal PID control and lateral Pure Pursuit. One scenario and results are shown in Fig. \ref{fig:fieldtest}. As in Table \ref{tab:field}, CSDO achieves similar solution quality while the runtime is greatly reduced.

\section{Conclusion}
This work introduces CSDO, an efficient algorithm for large-scale multi-vehicle trajectory planning, leveraging a combination of centralized priority-based searching and decentralized optimization. 
Through an extensive set of experiments, we demonstrate that CSDO efficiently discovers solutions within a limited time compared to other methods, without significant loss in solution quality, especially in large-scale, high-density scenarios. In the future, we will try to strike a better balance between solution quality and runtime with SOTA MAPF algorithms, generate robust solutions allowing for tracking errors, and extend our algorithm with dynamic and intensive traffic participants.






\vfill

\end{document}